\documentclass[lettersize,journal]{IEEEtran}

\usepackage[utf8]{inputenc}
\usepackage{lipsum}
\usepackage{multicol}
\usepackage{graphicx}
\usepackage{enumerate}
\usepackage{enumitem}
\usepackage{romannum}
\usepackage{amsmath}
\usepackage{amssymb}
\usepackage{multirow}
\usepackage{tabularray}
\usepackage[hyphens]{url}
\usepackage{hyperref}
\hypersetup{breaklinks=true}
\usepackage{makecell}
\usepackage{cite}
\usepackage{algorithm}
\usepackage{algpseudocode}
\usepackage{caption}
\usepackage{subcaption}
\usepackage{xcolor}
\usepackage[normalem]{ulem}

\newcommand{\add}[1]{\textcolor{black}{#1}}
\newcommand{\addfinal}[1]{\textcolor{black}{#1}}

\newcommand{\denote}[2]{$\textbf{\textit{#1}}_\text{#2}$}
\newcommand{\hatdenote}[2]{$\hat{\textbf{\textit{#1}}_\text{#2}}$}
\newcommand{\tildenote}[2]{$\Tilde{\textbf{\textit{#1}}_\text{#2}}$}
\newcommand{\vecdenote}[2]{$\Vec{\textbf{\textit{#1}}}_{\textit{#2}}$}
\newcommand{\vectildenote}[2]{$\Vec{\Tilde{\textbf{\textit{#1}}}}_{\textit{#2}}$}

\title{
\addfinal{Forbidden Region Dynamic Active Constraints in Robot-Assisted Minimally Invasive Surgery}}

\author{Zejian Cui$^{1}$ and Ferdinando Rodriguez y Baena$^{1}$

\thanks{\addfinal{Manuscript received: 7 August 2024; Revised 4 November 2024; Accepted 14 January 2025.}}
\thanks{\addfinal{This paper was recommended for publication by Editor Jessica Burgner-Kahrs upon evaluation of the reviewers' comments.
This work was supported by scholarship from the Department of Mechanical Engineering, Imperial College London.} \textit{(Corresponding author: Zejian Cui)}} 

\thanks{$^{1}$Both authors are with the Mechatronics in Medicine Laboratory, the Hamlyn Centre for Robotics Surgery, Department of Mechanical Engineering,
        Imperial College London, Exhibition Road, London, SW7 2AZ, UK
        {\tt \footnotesize \{zejian.cui19, f.rodriguez\}@imperial.ac.uk}}
\thanks{\addfinal{Digital Object Identifier (DOI): see top of this page.}}

}

\graphicspath{{figures/}}

\begin{document}

\markboth{\addfinal{IEEE Robotics and Automation Letters. Preprint Version. Accepted January, 2025}}
{Cui \MakeLowercase{\textit{et al.}}: FRDAC} 

\maketitle

\begin{abstract}
    In robot-assisted surgery, Forbidden Region Active Constraints (FRAC) represent a control strategy that helps maintain task safety by generating anisotropic haptic guidance to surgeons. However, several challenges need to be overcome before FRAC can benefit teleoperative surgery in a clinical setting. These challenges include the ability to allow for dynamic tissue deformation, maintain energetic passivity, and speed of implementation, among others. In this study, we propose the pipeline design for an energy dissipative FRAC strategy, which accommodates the dynamic tissue deformation caused by respiratory movements, by utilizing a depth sensing camera. The proposed FRAC strategy adopts a fine mesh representation, with a total number of 122,806 polygons in the case study presented, while running at 43.48Hz. We designed \textit{in vitro} trajectory tracking experiments conducted by a ``virtual" surgeon to aid quantitative assessment of the method, including its effectiveness in maintaining task safety, which was confirmed by successfully maintaining a pre-defined safety distance across all trials. We also conducted comparative studies to investigate the robustness and time-efficiency of our method against other FRAC methods that rely on simple geometry AC representations. We demonstrate that our method provides a more robust and effective guidance overall, while maintaining comparable, if not lower, time costs.
\end{abstract}

\begin{IEEEkeywords}
\addfinal{Medical Robots and Systems, Vision-Based Navigation, Motion Control.}
\end{IEEEkeywords}
\section*{List of acronyms}
\addcontentsline{toc}{section}{Nomenclature}
\begin{IEEEdescription}[\IEEEusemathlabelsep\IEEEsetlabelwidth{$V_1,V_2,V_3$}]
\item[\textbf{AC}] Active Constraints.
\item[\textbf{CTG}] Constrained Tool Geometry.
\item[\textbf{DAC}] Dynamic Active Constraints.
\item[\textbf{DIBS}] Deformation Invariant Boundary Spheres.
\item[\textbf{FRDAC}] Forbidden Region Dynamic Active Constraints.
\item[\textbf{FPFH}] Fast Point Feature Histograms.
\item[\textbf{HIP}] Haptic Interaction Point.
\item[\textbf{MIS}] Minimally Invasive Surgery.
\item[\textbf{PSM}] Patient Side Manipulator.
\end{IEEEdescription}
\section{INTRODUCTION}

\IEEEPARstart{I}{n} the field of robot-assisted surgery, Active Constraints (AC), also known as Virtual Fixtures (VF), were first proposed by Rosenberg \cite{rosenberg_use_1992,Rosenberg_1993} as strategies to ensure surgical task safety by providing surgeons with anisotropic haptic guidance. AC allow motions that comply with safety requirements, while negating those that violate safety requirements. Clinical practice has confirmed the effectiveness of AC in reducing the mental workload of surgeons and enhancing task performance \cite{AC_Acrobot}, where AC were employed to assist a hands-on knee replacement operation. Additionally, AC can be helpful in teleoperative surgical scenarios, where the physical separation between the patient side and the surgeon side prevents surgeons from forming haptic awareness of a surgical instrument during operation, as they would in hands-on surgery. The lack of haptic information forces surgeons into evaluating task safety purely through visual feedback, which poses a cognitive challenge to inexperienced surgeons \cite{daVinci_haptic_limitation}.


AC can be divided into two categories based on their application: FRAC, which help prevent potential violations of the safety region and tool collision, and Guidance AC (GAC), which help keep surgical instruments on a pre-planned trajectory or move them towards a target. In Robot-Assisted Minimally Invasive Surgery (RMIS), our focus is on task safety, and we are specifically concerned with FRAC. There are several challenges that need to be overcome before FRAC can be incorporated into a modern clinical setting. First, to provide surgeons with smooth haptic guidance, the speed at which haptic cues are sent needs to reach at least 1,000 Hz \cite{1000HZ_requirement}. Second, during operation, soft tissues undergo deformation caused by respiration and tool-tissue interactions, requiring FRAC to be updated in real-time to adjust to a dynamic situation. Third, the accuracy requirement in RMIS is more stringent, posing a greater challenge for the implementation of FRAC. For example, the cauterization task in kidney cancer surgery requires maintaining a safety distance of 5 mm between the laser fiber held by a robot end effector and the kidney surface \cite{Ryden_journal}.

To address these challenges, prior work has focused on developing Forbidden Region Dynamic Active Constraints (FRDAC), which provides more advantages in incorporating real-time tissue changes compared to FRAC. Although there exists studies such as \cite{marinho2019dynamic}, where collision avoidance between dynamic objects is achieved through monitoring distance functions and distance Jacobians, which encode dynamic factors, relying purely on the robot data, a more common practice is to incorporate an external sensing modality. Ren \emph{et al.} \cite{ren2008dynamic} first proposed a FRDAC pipeline for beating heart surgery, where preoperative CT/MRI images are continuously registered to the real-time ultrasound images, enabling the update of FRDAC on the surface of a beating heart. Using a different imaging modality as a stereo camera, Yamamoto \emph{et al.} \cite{yamamoto2012augmented} incorporated in their haptic interfaces reconstruction of surface point cloud, upon which FRAC is constructed. More directly,  Ryd{\'e}n \emph{et al.} 
 \cite{Ryden_proxy_rudimentary,Ryden_proxy} used a depth-sensing camera, Xbox Kinect (Microsoft Corp.), in their FRDAC design to protect a beating heart. The camera streams a point cloud representing the surface of a beating heart at 30Hz, so that the AC constructed around each point on the tissue surface could be constantly updated.

One existing issue among the FRDAC methods above is that the accuracy requirement can conflict with the objective of targeting real-time applications in FRDAC design. Ryd{\'e}n \emph{et al.} \cite{Ryden_proxy} chose simple geometries, specifically spheres, to represent AC. This reduces the time required for each round of AC implementation, making it more suitable for clinical application. However, simple geometries are unable to represent complex tissue surface features \cite{Stuart_review}, which can reduce the accuracy of AC implementation. On the other hand, representing FRDAC using complex geometries always involves conducting nonrigid registration from a finely described non-deformed tissue surface to a dynamically moving tissue surface. Non-rigid registration methods, such as the Coherent Point Drift (CPD) \cite{CPD} method, for example, can be computationally demanding \textit{per se}, which may be challenging when capturing the deformation caused by respiration at a rate of 2Hz. Another issue is that most of the existing FRDAC pipelines do not consider energy dissipativity. This is important because it ensures that there is no energy stored from user input, and hence precludes surgical tools being actively driven to move without the active involvement of surgeons, thus maintaining task safety and avoiding liability issues. Existing non-energy storing methods include controllers based on a plasticity model \cite{kikuuwe2008control}, an energy redirection approach based on an elasto-plastic friction model \cite{StuartDynamicTRO}, and controllers based on a viscosity model \cite{enayati2016dynamic}.

In this study we present a FRDAC pipeline designed to prevent soft tissue violations in a dynamic situation, while ensuring the entire system remains energetically dissipative. In this paper we:
\begin{itemize}
    \item Propose an approach to generating FRDAC that are capable of describing complex geometric features of a deforming tissue surface whilst satisfying the requirement for real time application.
    \item Incorporate a dissipative controller into the design of FRDAC which helps to guarantee overall safety during operation. 
    \item Conduct \textit{in vitro} experiments to verify the effectiveness of the proposed FRDAC implementation pipeline and quantify the impact of the complexity of AC representation on task performance in a dynamic scenario.
\end{itemize}

The structure of this paper is defined as follows. In section \ref{sec:methods}, we provide a detailed description of the proposed FRDAC implementation pipeline. In section \ref{sec:experiments}, we describe the experimental design of our study. The result and discussion are presented in section \ref{sec:results_discussion}, followed by our conclusion in section \ref{sec:conclusion}.

\section{Methods} \label{sec:methods}
According to \cite{Stuart_review}, a standardized pipeline for AC implementation consists of three interrelated stages, namely AC generation, AC evaluation and AC enforcement. In the remainder of this section, the description of our FRDAC design will be divided into three subsections, following the order of AC implementation.

\subsection{AC Generation}
\label{subsec:AC_generation}
AC generation is crucial to represent the tissue surface to be protected using constraints. The complexity of these geometrical representations affects the difficulty of checking for safety region violations, which occurs in the AC evaluation phase. To enable a more detailed representation of deforming tissue surface features, we use triangle meshes for AC generation instead of surface primitives, such as spheres and planes. To monitor tissue deformation and to enable online updates of AC, we incorporate a depth sensing camera into our pipeline design. The registration process is illustrated in Fig.\ref{fig_AC_generation}.

To update mesh-represented FRDAC via registering the pre-scanned tissue model to the point cloud describing the current state of the tissue surface, non rigid registration methods such as the CPD method \cite{CPD} can be used. Although the classic CPD method has been greatly improved to accelerate the speed of non-rigid registration, as shown in examples such as \cite{BCPD,BCPD++}, these model-free registration methods still fail to satisfy the real-time requirement.
To address this issue, We assume that the deformation caused by respiration can be modeled as an affine transformation.
Under this assumption, we incorporated the TEASER++ method \cite{Yang20tro-teaser} into our pipeline design , which enables rapid and robust global affine registration. To further enhance the robustness of the registration, we first extract the Fast Point Feature Histograms (FPFH) features \cite{rusu2009FPFH} from both the tissue model and the target point cloud to establish the initial correspondences, which are then leveraged by the TEASER++ method. 

To verify the assumption above, we first created two deformed aorta phantom models in Blender \cite{Blender}, reflecting the status of inspiration and expiration, by 
applying inhomogeneous displacement along the surface normal direction to vertices in different anatomical regions. The amount of the applied displacement is based on the study in \cite{aorta_quantification}. Then we compared the registration performance between TEASER++ and the non-rigid registration algorithm BCPD++ in \cite{BCPD++} for cases of both inspiration and expiration. The average registration error is defined as $\frac{1}{N}\sum_{i=1}^N||P_i^{gt}-P_i^{reg}||$, where N represents the total number of vertices on the phantom model, $P_i^{gt}$ represents the ground truth position of the $i^{th}$ vertex in the deformed model, and $P_i^{reg}$ represents the position of the $i^{th}$ vertex after registration. In the case of expiration, the average registration error using TEASER++ is 2.52 mm with a registration size of (8,000,8,000)\footnote[1]{\add{It represents the registration between 8,000 source points and 8,000 target points. The same definition applies to (2,000,2,000).}}, and the error using BCPD++ is 2.55 mm, with a registration size of (2,000,2,000). In the case of inspiration, the average error is 2.40 mm using TEASER++, and 2.34 mm using BCPD++.

To further speed up the registration process, we perform FPFH feature extraction only once in the first round of the FRDAC implementation, and we assume that the feature correspondences are preserved in the following rounds of implementation as long as the normal features of the tissue surface do not change drastically during operation. After registration, we can update points on the CAD to the camera frame, shown in Eq.\ref{equa:AC generation}.
\begin{equation}\label{equa:AC generation}
    \{P_{model}\} \xrightarrow[T_{4\times4}, scale]{\text{Affine registration}} \{P_{model}^*\}
\end{equation} 

This assumption is supported by our simulation results presented in Fig.\ref{fig:FPFH_analysis}. To verify the assumption, we first extracted FPFH feature correspondences between the non-deformed and inspiration model, and then used these correspondences for the registration to the expiration model. The registration test was compared against when FPFH feature correspondences are freshly established, across different sizes of point cloud. The t-tests results show that there is no statistical difference between the two approaches when the size of registration is not larger than 8,000.

\begin{figure}[t]
\centering
\includegraphics[width=\columnwidth]{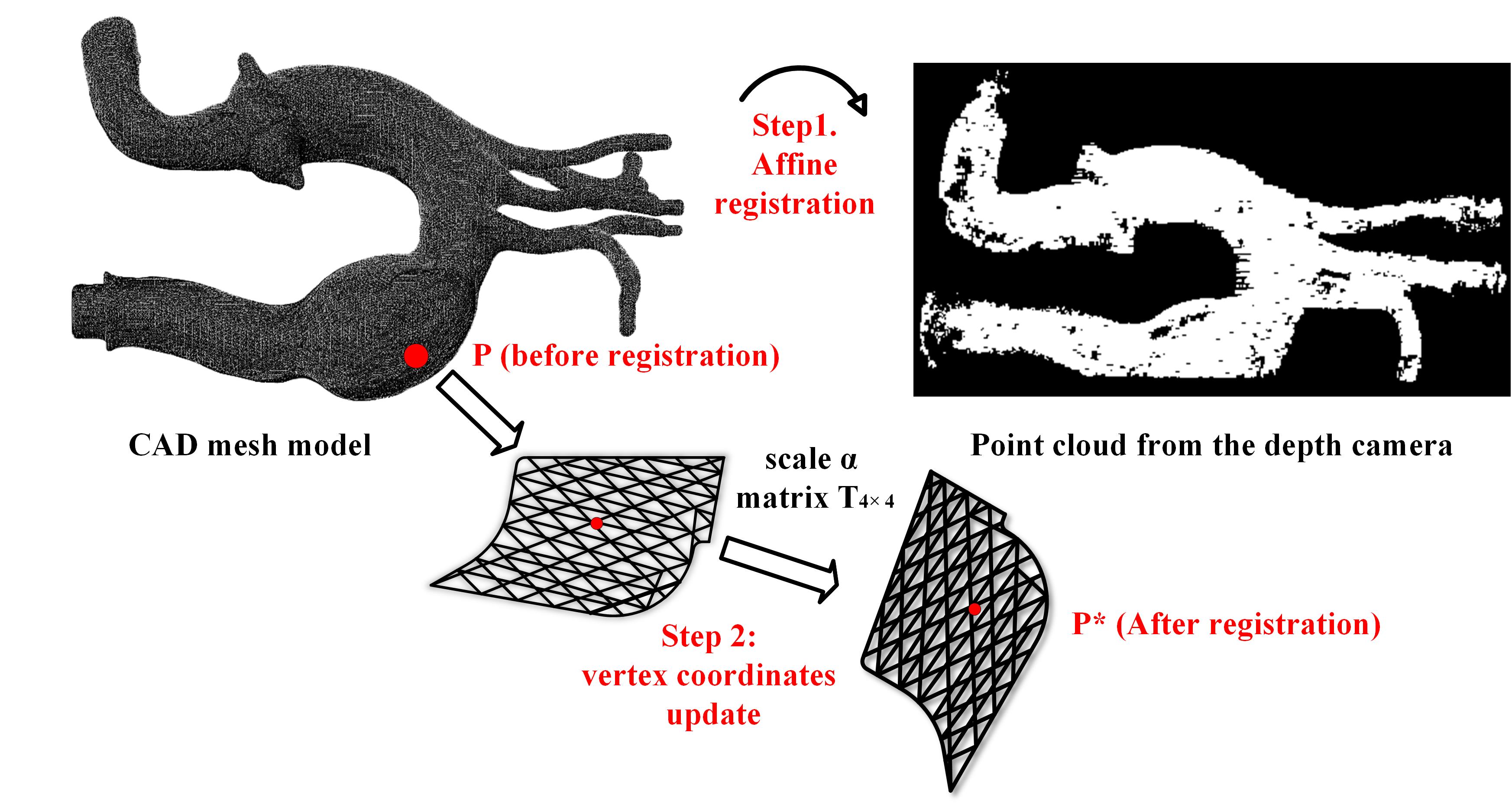}
\setcounter{figure}{0}
\caption{Mesh-represented FRDAC generation}
\label{fig_AC_generation}
\end{figure}

\subsection{AC Evaluation}
For FRDAC, AC evaluation means detecting whether there exists any violation of the safety region. During operation, the specific area on a surgical instrument that must be prevented from damaging tissue surfaces is defined as Constrained Tool Geometry (CTG), and a violation of the safety region occurs when \add{$d_{min}$}, the minimum distance between the CTG and the tissue surface, falls within a pre-defined safety distance \add{$d_{safe}$}. \add{$d_{min}$ is also the distance between the closest point pair, which is formed by $P_{tool}^*$ on the CTG, and $P_{tissue}^*$ on the tissue surface.} To accelerate the proximity query process, in this study, we adopt the Deformation Invariant Boundary Spheres (DIBS) method \cite{DIBS}, which enables the proximity query between a cylindrical CTG and a dynamically moving brain mesh model with a resolution of 500,000 polygons to run at 1,587 Hz without GPU acceleration. The AC evaluation procedure is summarised in Eq.\ref{equa:AC evaluation}.

\begin{equation}\label{equa:AC evaluation}
    \text{min } \text{distance}(\text{CTG}, P_{model}^*) \xrightarrow[d_{min}]{\text{DIBS query}} \langle P_{tool}^*, P_{tissue}^* \rangle
\end{equation}

\subsection{AC Enforcement}
\begin{figure}[t]
\centering
\includegraphics[width=\columnwidth]{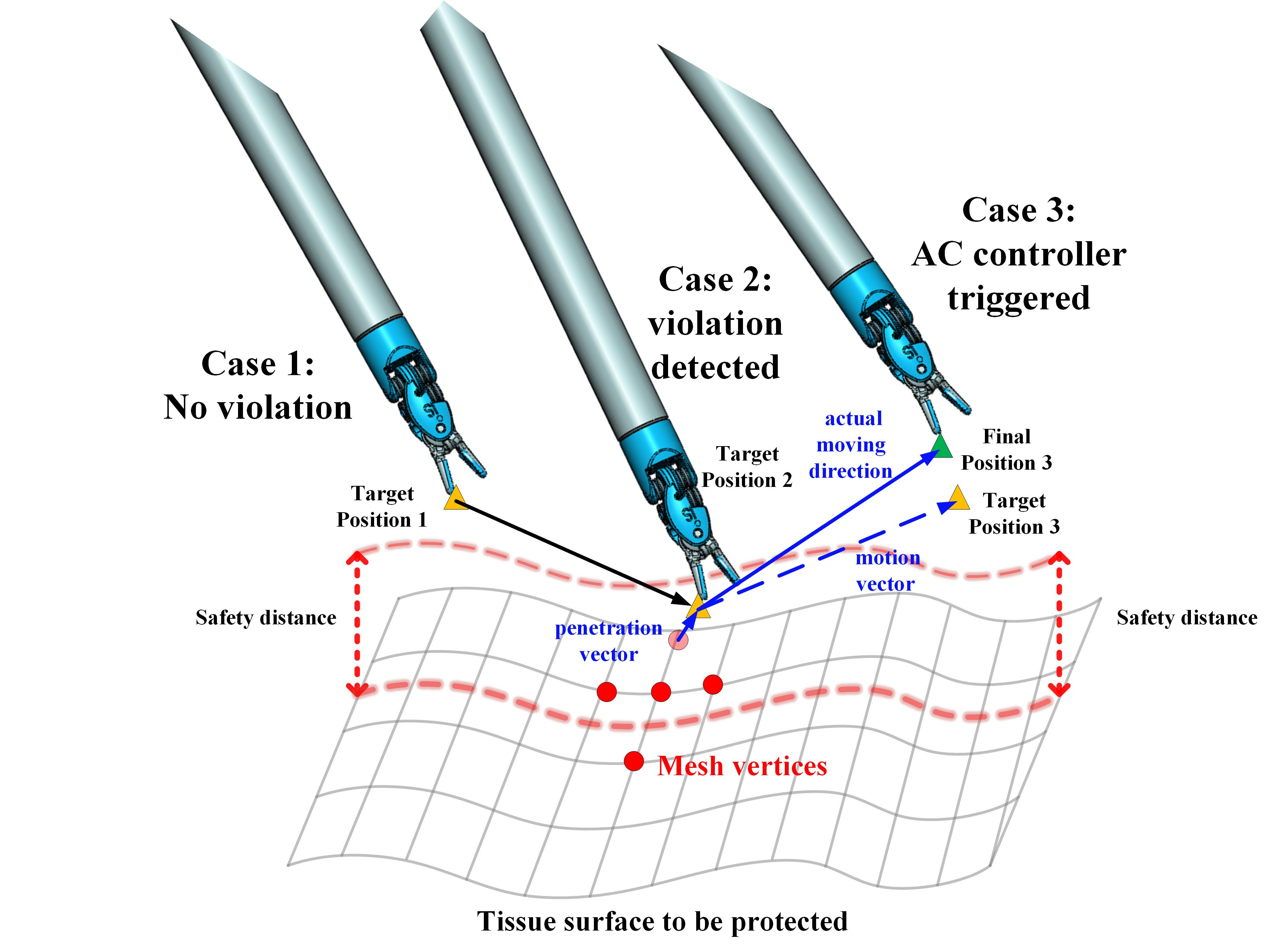}
\caption{Illustration of AC evaluation and AC enforcement. \add{$\Vec{\textit{v}}_{\textit{act}}$ is calculated as the elastic displacement $\textbf{z}_\textbf{t}$ using the energy redirection approach in \cite{Stuart_Enforcement_ICRA}.}}

\label{fig_AC_evaluation}
\end{figure}

The final stage of AC implementation is to compute the magnitude and the direction of the AC force to be sensed by a surgeon such that, when a violation of the safety region is detected, the surgeon can promptly retract the surgical instrument away from the moving tissue. In this study, we adopt the dissipative controller \cite{Stuart_Enforcement_ICRA} enabled by energy redirection for AC force generation to maintain the dissipativity of the entire system. We refer to the unit vector that connects the \add{closest point pair as penetration vector ($\Vec{\textit{v}}_{\textit{pen}}$)}, and the unit vector that aligns with the direction of the current needle tip movement as motion vector ($\Vec{\textit{v}}_{\textit{mov}}$), pointing \add{from the current tip point $P_{tip}$} towards the target point $P_{tgt}$. After sending $\Vec{\textit{v}}_{\textit{pen}}$ and $\Vec{\textit{v}}_{\textit{mov}}$ into the dissipative controller, which works in $\mathbb{R}^{3}$, the friction component $\Vec{\textit{v}}_{\textit{act}}$ will be calculated, \add{as detailed in \cite{Stuart_Enforcement_ICRA}}, which is then utilised to generate an AC force $\Vec{F_{\textit{AC}}}$ \add{based on an elastic-plastic model}. A graphical illustration is shown in Fig.\ref{fig_AC_evaluation}. Eq \ref{eq:AC_enforcement} summarizes the AC evaluation process, where $\Sigma_0, \Sigma_1$ represents a $3\times3$ friction stiffness matrix and viscous coefficient matrix, respectively.

\begin{equation}\label{eq:AC_enforcement}
\begin{split}
        \Vec{\textit{v}}_{\textit{pen}}&=(P_{tool}^*-P_{tissue}^*)/d_{min} \\
    \Vec{\textit{v}}_{\textit{mov}}&=(P_{tgt}-P_{tip})/||(P_{tgt}-P_{tip}|| \\
    \Vec{F_{\textit{AC}}}&=\Sigma_0 \Vec{\textit{v}}_{\textit{act}} + \Sigma_1 \dot{\Vec{\textit{v}}}_{\textit{act}}
\end{split}
\end{equation}
\section{Experiments} \label{sec:experiments}

\subsection{Overview}
To investigate the effectiveness of the proposed FRDAC pipeline in maintaining task safety in MIS, we conducted experiments where the proposed FRDAC assisted a surgical instrument in performing a trajectory tracking task whilst maintaining a preset safety distance $d_{safe}$. As a proof of concept implementation, the trajectory tracking experiments were conducted \textit{in vitro}. To remove the human element from this study, we incorporated a hybrid force-position controller to emulate the ideal situation where a ``virtual" surgeon could perfectly act upon the haptic force generated by the AC controller. Additionally, we compared our FRDAC against the FRDAC developed in \cite{Ryden_proxy} 
in assisting the same trajectory tracking task to quantify the effect of AC representation on the efficiency of AC implementation. We designed experiments in which both methods were used for assisting with trajectory tracking tasks in both static and dynamic scenarios, across various levels of deformation. The remainder of this section describes the hardware setup, the design of the hybrid force-position controller, and the algorithm designed for the comparative experiments.

\begin{figure}[t]
\centering
\includegraphics[scale=0.42]{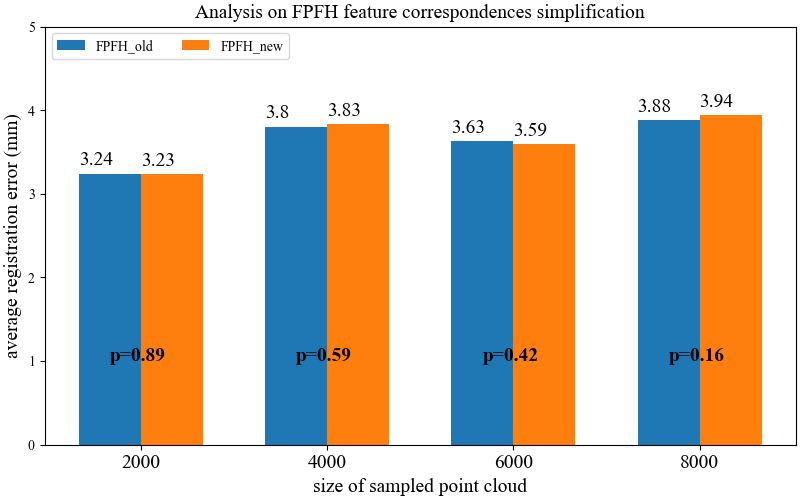}
\setcounter{figure}{1}
\caption{Comparison of average point registration errors across different size of registration. \text{``FPFH\_new''} refers to when correspondences are freshly established; \text{``FPFH\_old''} refers to when correspondences are assumed unchanged. Altogether, there are 12,000 vertices on the phantom model. The p value of t tests are also presented.}
\label{fig:FPFH_analysis}
\end{figure}

\begin{figure}[t]
\centering
\includegraphics[scale=0.22]{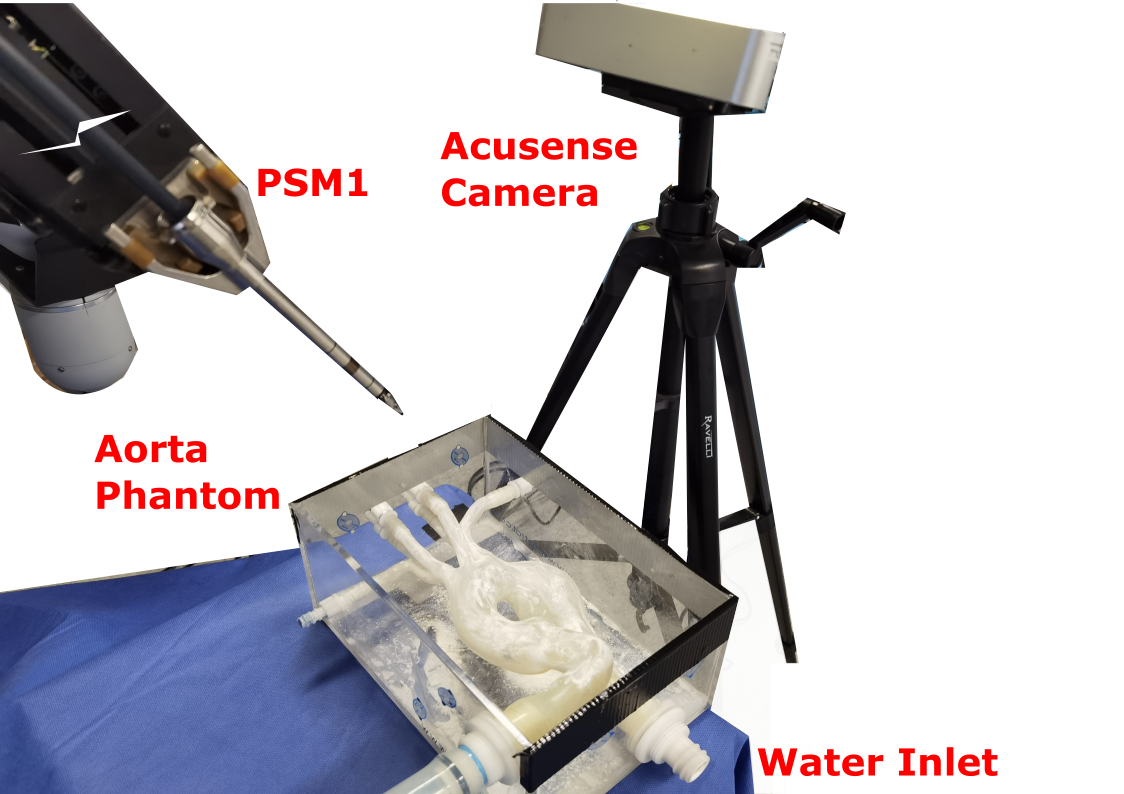}
\caption{Hardware setup, consisting of an Acusense 3D camera, an aorta silicone phantom and a pair of forceps held by dVRK Patient Side Manipulator1 (PSM1)}

\label{fig_example}
\end{figure}

\subsection{Hardware Setup}
The experiments in this study were conducted using the first-generation da Vinci Research Kit (dVRK) \cite{dVRK_2014}. A surgical instrument (Forceps 400036E, Intuitive Inc$^{\circledR}$) attached to Patient Side Manipulator1 (PSM1) tracked a dynamic trajectory on the surface of a deforming aorta silicone phantom that was controlled by a water pump. We used a RGBD camera, the Acusense 3D (Revopoint 3D Inc$^{\circledR}$), with a resolution of 0.5 mm at a working distance of 500 mm, and a depth capturing rate of 10 Hz. The algorithm for FRDAC implementation was written in C++ and executed as a ROS node on Ubuntu 20.04 installed on a desktop (EliteDesk 800 G5, HP corp.) without GPU acceleration. 

\subsection{Hybrid Force-Position Controller}


We postulate that, when an AC force is generated, it can effectively slide the surgical tool in the direction of the AC force as if a ``virtual" surgeon was controlling it. To facilitate such movement, we introduce a hybrid force-position controller, based on the approach presented in \cite{hybrid_force_position1981}, which computes the displacement of the end effector of the surgical tool in Cartesian space as: $ \Vec{\Delta x} = [\mathcal{S}]\Vec{F_{\textit{AC}}}$,
where $\Vec{\Delta x}$ is a \add{$3\times1$} vector denoting the \add{translational} displacement of the end effector of the surgical tool in Cartesian space; \add{$[\mathcal{S}] = \text{diag}\{s_1,s_2,s_3\}$} denotes a \add{$3\times3$} compliance matrix, and $\Vec{F_{\textit{AC}}}$ denotes a \add{$3\times1$} vector of the generated AC force. \add{In this way, the direction of $\Vec{\Delta x}$ changes as $\Vec{F_{\textit{AC}}}$ varies after each round of AC enforcement.}

\subsection{Algorithm Implementation}
\subsubsection{Preparation tasks}
A reference trajectory was predefined on the non-deformed phantom surface. Prior to the experiments, the Acusense camera was required to be positioned at a distance between $500$ mm and $600$ mm from the phantom surface, and the transformation matrix from the camera frame to the PSM1 base frame was required. \add{Detailed procedures are described below.}

\begin{enumerate}[label=\roman*]
    \item Reference Trajectory Generation:
    \\ We designed two circular reference trajectories \denote{Ref}{1} and \denote{Ref}{2} for static and dynamic experiments, respectively. Both trajectories are generated by first manually selecting $\textit{N}_{\textit{raw}}$ vertex points located on different anatomical regions of the phantom mesh model, and then finding the best-fit circle that contains $\textit{N}_{\textit{full}}$ points using the circle-fitting-3d library \footnote[2]{\url{https://pypi.org/project/circle-fitting-3d/}}. Then we translated the best-fit circle along the normal direction of the circular plane by $d_n$ to generate reference trajectories within varying distances from the phantom surface. These trajectories travel through different regions of the aorta while providing easy access to testing both FRDAC strategies in situations where safety regions are violated. We chose $d_n=1$ cm, $\textit{N}_{\textit{full}}=72$ for \denote{Ref}{1}, and $d_n=2$ cm, $\textit{N}_{\textit{full}}=72$ for \denote{Ref}{2}.
    \item Hand-eye Calibration:
    \\ The transformation matrix ($\textit{T}_{cr}$) from the camera base frame to the PSM1 base frame was obtained from preoperative hand-eye calibration using two 3D-printed spheres, following the process described in \cite{Berkeley_Automation}. Altogether, we gathered $50$ point pairs to calculate the transformation matrix.  
\end{enumerate}

\subsubsection{Workflow}
In the comparison study, experimental trials that adopted the FRDAC developed in this study are denoted as \denote{Exp}{ours}, and the ones that adopted the FRDAC proposed in \cite{Ryden_proxy} are denoted as \denote{Exp}{Ryd{\'e}n}. The differences in the two implementations lie in how the FRDAC is updated, and how to determine the desired tool position when a violation of the safety region is detected. In \denote{Exp}{Ryd{\'e}n}, the FRDAC is constructed as spheres around the points on the tissue surface, and the desired tool tip position (proxy position) keeps being updated from the current position, also called Haptic Interaction Point (HIP), until the proxy state is free. 

The workflows using different FRDAC methods are described as follows and summarized in Algorithm.\ref{alg:dataflow}. After launching the ROS node, the camera first streams the point cloud on the tissue surface \hatdenote{Cloud}{tissue}, to which the point cloud on the non-deformed tissue model \denote{Cloud}{tissue} is registered, returning the affine transformation matrix $\textit{A}_{mc}$ from the model frame to the camera frame. To accelerate the registration process, \hatdenote{Cloud}{tissue} is uniformly downsampled to $\textit{N}_{m}$ points, and \denote{Cloud}{tissue} is uniformly downsampled to $\textit{N}_{s}$ points. With $\textit{A}_{mc}$, the reference trajectory in the model frame \denote{Ref}{} is then updated to the camera frame as \hatdenote{Ref}{}. Subsequently, \hatdenote{Ref}{} and \hatdenote{Cloud}{tissue} are transformed to the PSM1 base frame as \tildenote{Ref}{} and \tildenote{Cloud}{tissue}, respectively, through $\textit{T}_{cr}$. We then read the joint values of PSM1 from the dVRK API to obtain the tip position \tildenote{P}{tip}. The tool tip is required to cover all the points on \tildenote{Ref}{} progressively from the first one, and when $||$\tildenote{P}{tar}$-$\tildenote{P}{tip}$||<\epsilon$, the target point is updated by one.


To ensure  fairness of the comparison, the same dissipative controller is adopted in \denote{Exp}{ours} and \denote{Exp}{Ryd{\'e}n}, but with different definitions in \vecdenote{v}{pen}. In \denote{Exp}{ours}, \vecdenote{v}{pen} is defined \add{as in subsection \ref{subsec:AC_generation}}, whereas in \denote{Exp}{Ryd{\'e}n}, \vecdenote{v}{pen} is defined by the unit vector connecting the HIP \add{(instrument tip)} to the updated proxy point \add{(desired tip position on the forbidden region surface)}. In both cases, \vecdenote{v}{mov} is defined as the unit vector pointing from \tildenote{P}{tip} to \tildenote{P}{tar}. Both \vecdenote{v}{pen} and \vecdenote{v}{mov} are then sent to the AC controller to generate a frictional force \vectildenote{f}{AC}. Eventually, \vectildenote{f}{AC} is sent to the hybrid force-position controller to generate a positional command \vectildenote{$\Delta x$}{AC} to be received by PSM1.

\begin{algorithm}
\caption{Workflow for Dynamic AC}\label{alg:dataflow}
\begin{algorithmic}
    \Require \denote{Cloud}{tissue}, $\textit{T}_{cr}$
    \State \textbf{int} $i \gets 1$ \Comment{index for the current target point}
    \State \textbf{bool} $\textit{violation} \gets \textit{false}$
    \While{$i \leq \textit{N}_{\textit{full}}$}
    \State \hatdenote{Cloud}{tissue} $\gets$ \hatdenote{Cloud}{tissue} \Comment{Acusense streaming}
    \State $\textit{A}_{mc} \gets \textit{A}_{mc}$ \Comment{TEASER++}
    \State \textit{update} \denote{Ref}{} \textit{to} \hatdenote{Ref}{}, \tildenote{Ref}{}
    \State \tildenote{P}{tar} $\gets$ \tildenote{Ref}{}$[i]$
    \If{$||$\tildenote{P}{tar}$-$\tildenote{P}{tip}$||<\epsilon$}
        \State $i \gets min \langle i+1 , \textit{N}_{\textit{full}} \rangle$
        \State \tildenote{P}{tar} $\gets$ \tildenote{Ref}{}$[i]$
    \EndIf
    \If{\denote{Exp}{ours} is on} \Comment{mesh-represented FRDAC}
        \State \textit{update} $\{$\denote{point}{tissue}, \denote{point}{tool}, \denote{d}{min}$\}$
        \If{$\textit{d}_{\textit{min}} \leq d_{safe}$}
            \State $\textit{violation} \gets true$
            \State \vecdenote{v}{pen} $\gets$ $\frac{\text{\denote{point}{tissue}}-\text{\denote{point}{tool}}}{||\text{\denote{point}{tissue}}-\text{\denote{point}{tool}}||}$
            \State \vecdenote{v}{mov} $\gets$ $\frac{\text{\tildenote{P}{tar}}-\text{\tildenote{P}{tip}}}{||\text{\tildenote{P}{tar}}-\text{\tildenote{P}{tip}}||}$
        \EndIf
    \ElsIf{\denote{Exp}{Ryd{\'e}n} is on} \\ \Comment{sphere-represented FRDAC}
        \State \denote{HIP}{} $\gets$ \tildenote{P}{tip} 
        \State \denote{proxy}{} $\gets$ \denote{proxy}{updated}
        \If{\denote{HIP}{}$ \neq $\denote{proxy}{}}
            \State $\textit{violation} \gets true$
            \State \vecdenote{v}{pen} $\gets$ $\frac{\text{\denote{proxy}{}}-\text{\denote{HIP}{}}}{||\text{\denote{proxy}{}}-\text{\denote{HIP}{}}||}$
            \State \vecdenote{v}{mov} $\gets$ $\frac{\text{\tildenote{P}{tar}}-\text{\tildenote{P}{tip}}}{||\text{\tildenote{P}{tar}}-\text{\tildenote{P}{tip}}||}$
        \EndIf
    \EndIf
    \If{$\textit{violation}=\textit{true}$}
        \State \vectildenote{f}{AC} $\gets$ $\textit{AC}_{\textit{Enforce}}(\text{\vecdenote{v}{pen}}, \text{\vecdenote{v}{mov}}, \text{\tildenote{P}{tip}})$
        \State \vectildenote{$\Delta x$}{AC} $\gets$ $\textit{Hybrid}( \text{\vectildenote{f}{AC}} )$
        \State \vectildenote{$\Delta x$}{} $\gets$ $\frac{\text{\vectildenote{$\Delta x$}{AC}}}{||\text{\vectildenote{$\Delta x$}{AC}}||}$
    \Else
        \State \vectildenote{$\Delta x$}{} $\gets$ $\frac{\text{\tildenote{P}{tar}}-\text{\tildenote{P}{tip}}}{||\text{\tildenote{P}{tar}}-\text{\tildenote{P}{tip}}||}\textit{min}\langle||\text{\tildenote{P}{tar}}-\text{\tildenote{P}{tip}}||,\textit{stride}_{\textit{max}} \rangle$
    \EndIf
        \State $\textit{send}$ \vectildenote{$\Delta x$}{} $\textit{to }\textit{PSM}_\text{1}$
    \EndWhile
\end{algorithmic}
\end{algorithm}

\subsection{Deformation measurements}
To generate varying levels of deformation, we adjust the valve of the water pump connected to the phantom. To quantify the magnitude of deformation, we used the Acusense camera to capture a series of point clouds of the pulsating phantom surface within a given amount of time. The magnitude of deformation is calculated in Open3D\cite{O3D} as the range of distances between phantom surfaces in Open3D. We classify the levels of deformation into ``slow", ``fast" and ``super fast", where the magnitudes of deformation are $2.43$ mm, $3.64$ mm and $4.80$ mm, respectively. From visual observations, respiratory rate was set at around 1-2 Hz.


\subsection{Parameters selection}
In \denote{Exp}{ours}, we selected $N_m=8000; N_s=8000$. \add{For FPFH feature detection, FPFH search radius = $0.02$m, and FPFH normal radius = $0.06$m.} \add{For DIBS initialization, stretch factor = $1.2$ to allow for tissue deformation without the need of DIBS reconstruction}. Radius of cylindrical CTG = $0.005$m. $d_{safe}=0.010$m. In \denote{Exp}{Ryd{\'e}n}, \add{to achieve the optimal performance,} we \add{empirically} selected $r_1=0.00499$m, $r_2=0.00501$m, $r_3=0.01$m, $r_f=0.008$m, $\theta=1^{\circ}$, stiffness $K=600N/m$, and $d_{safe}=0.008$m. The parameters for the AC controller adopted in both \denote{Exp}{ours} and \denote{Exp}{Ryd{\'e}n} are listed as follows: $\{s_1, s_2, s_3\}$ in $\mathcal{S}$ takes $1000$; friction stiffness \add{in $\Sigma_0$} = $0.7N/mm$; \add{all} viscous friction coefficients \add{in $\Sigma_1$} = $0.01Ns/mm$; coulomb friction = $10.5N$; transition region = $0.002m$; minimum motion angle = $80^{\circ}$; maximum motion angle = $110^{\circ}$. $\epsilon=0.008$m in static experiments, and $\epsilon=0.015$m in dynamic experiments.

\section{Results and Discussion} \label{sec:results_discussion}
\subsection{Static trajectory tracking results} \label{subsec:static_tracking}
\begin{figure}[t]
\centering
\includegraphics[scale=0.22]{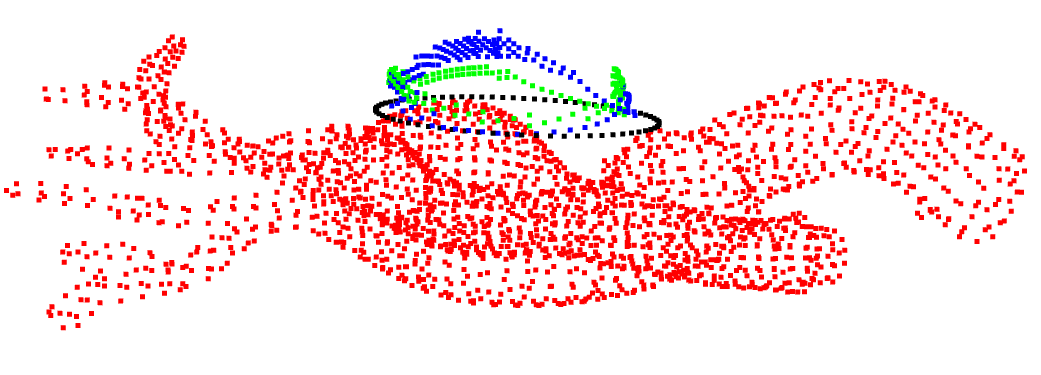}
\caption{Comparison of static trajectory tracking between \denote{Exp}{ours} and \denote{Exp}{Ryd{\'e}n}. \denote{Ref}{1} was used. Red represents the phantom model; black represents the reference circular trajectory \denote{Ref}{1}; green represents the actual trajectory in \denote{Exp}{ours}, and blue represents the actual trajectory in \denote{Exp}{Ryd\'en}.}
\label{fig:Static trajectory}
\end{figure}

\begin{figure}[t]
\centering
\includegraphics[height=4cm]{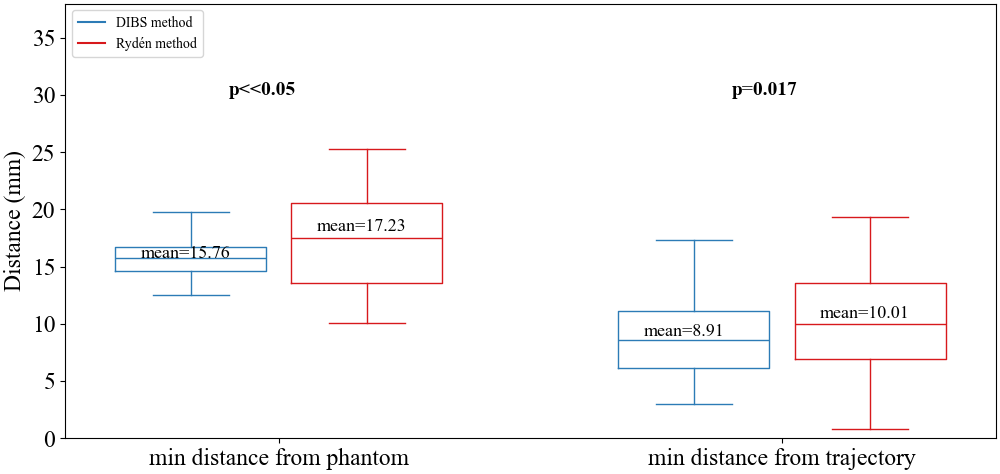}
\caption{Static trajectory tracking results. The minimum distance between the instrument and the moving phantom surface, as well as the constantly updated reference trajectory during \denote{Exp}{ours} and \denote{Exp}{Ryd{\'e}n} are presented. t tests were conducted \add{across a single trial}, and we denote $p<<0.05$ when $p<0.005$. Outliers are not shown in the figure, as in Fig.\ref{fig:phantom ttest}, Fig.\ref{fig:Traj ttest}, \add{and Fig.\ref{fig:AC_evaluation_time}.}}
\label{fig:static_tracking}
\end{figure}

For static trajectory tracking, we conducted only one trial using both FRDAC strategies for tracking \denote{Ref}{1}. \add{\denote{Ref}{1} and static trajectory tracking results are illustrated in Fig.\ref{fig:Static trajectory}.} The static trajectory tracking results are \add{plotted using matplotlib \cite{Matplotlib} in Python3.9 and} presented in Fig.\ref{fig:static_tracking}. 

In \denote{Exp}{ours}, a total of $149$ points were gathered on the actual tool tip trajectory. The average distance between the tool tip and updated target point on the reference trajectory $d_{traj}=\frac{1}{N}\sum_{i=1}^{N}||P_i - Q_i ||=8.91$ mm. where $P_i$ and $Q_i$ denote the $i^{th}$ point on the actual tool tip trajectory and its corresponding target point on the updated reference trajectory, respectively. The average minimum distance from the phantom surface $d_{phantom}=\frac{1}{N}\sum_{i=1}^{N}$\denote{d}{min,i}$=10.44$ mm, where \denote{d}{min,i} denotes the minimum distance returned by the the DIBS query at the $i^{th}$ instance.

In \denote{Exp}{Ryd{\'e}n}, a total of $170$ points were collected, and $d_{traj}$ was $10.01$ mm. $d_{phantom}=\frac{1}{N}\sum_{i=1}^{N}\textit{min}\langle P_i$, \denote{Cloud}{tissue,i} $\rangle$, where \denote{Cloud}{tissue,i} represents the downsampled tissue point cloud at the $i^{th}$ instance. $d_{phantom}$ in both \denote{Exp}{ours} and \denote{Exp}{Ryd{\'e}n} being greater than $d_{safe}=10$ mm indicates that both FRDAC methods can maintain task safety. Additionally, both $d_{traj}$ and $d_{phantom}$ in \denote{Exp}{ours} being smaller than those in \denote{Exp}{Ryd{\'e}n} suggests that the DIBS method favors accuracy over a larger safety margin.

\subsection{Dynamic trajectory tracking results}
\begin{figure}[t]
\centering
\includegraphics[height=4.5cm]{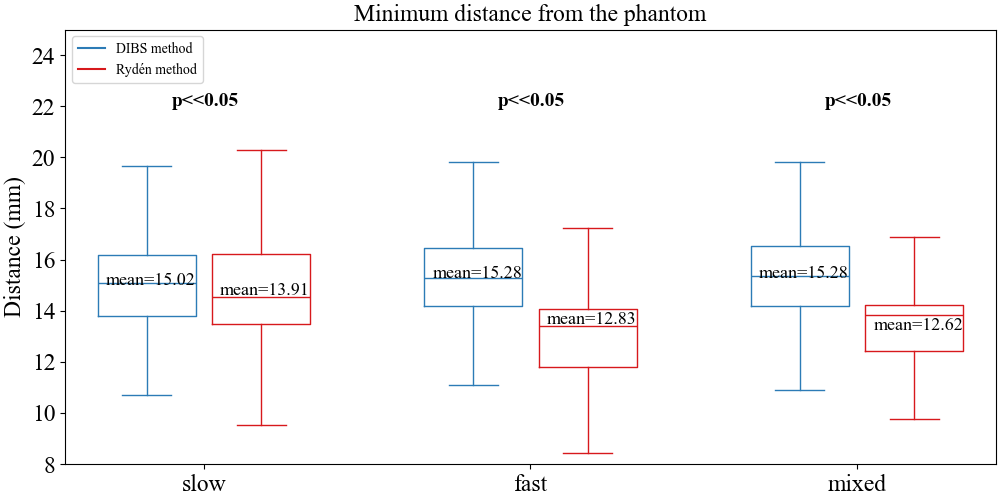}
\caption{Comparison of the minimum distance between the instrument and the moving phantom surface ($d_{phantom}$) between \denote{Exp}{ours} and \denote{Exp}{Ryd{\'e}n} across various levels of deformation.}
\label{fig:phantom ttest}
\end{figure}

\begin{figure}[t]
\centering
\includegraphics[height=4.5cm]{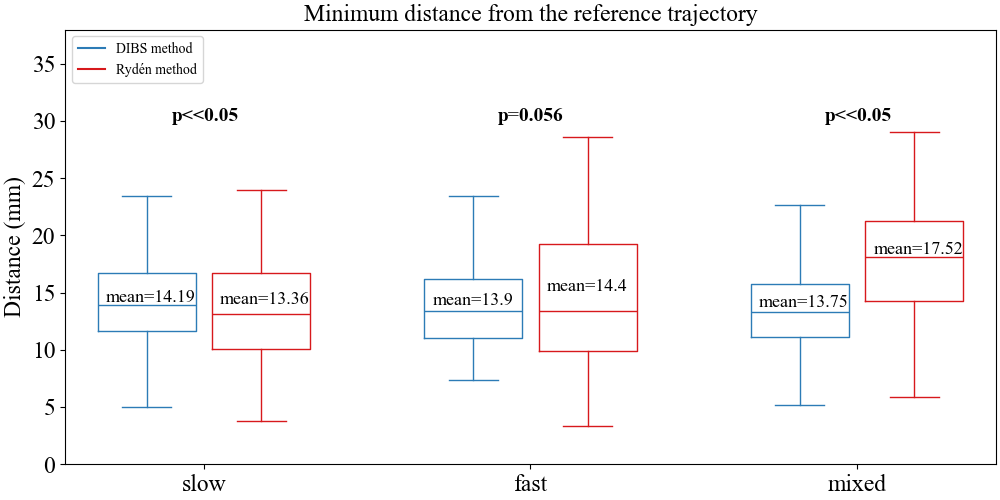}
\caption{Comparison of the minimum distance between the instrument and the constantly updated reference trajectory ($d_{traj}$) between \denote{Exp}{ours} and \denote{Exp}{Ryd{\'e}n} across various levels of deformation.}
\label{fig:Traj ttest}
\end{figure}

For dynamic trajectory tracking experiments, we conducted experiments across different levels of deformation, namely, ``slow", ``fast" and ``mixed" (where the level of deformation \add{changes} from ``slow" to ``fast" to ``super fast"). For each level of deformation, tracking experiments were conducted four times. We applied the same metrics as in \ref{subsec:static_tracking} to evaluate their tracking performance. \add{Dynamic trajectory of ``mixed run 4" is illustrated in Fig.\ref{fig:Dynamic Tracking, mixed run4}.} Tracking results using both FRDAC methods are presented in Fig.\ref{fig:phantom ttest} and Fig.\ref{fig:Traj ttest}, and the time costs for all the experimental trials are recorded in TABLE \ref{tab:time costs DIBS} and \ref{tab:time costs Ryden}.

From Fig.\ref{fig:phantom ttest}, it is noticed that in \denote{Exp}{ours} $d_{phantom} \geq d_{safe}=15$ mm for all the experimental runs, which indicates that the DIBS method can maintain task safety across a range of deformation levels. It has been noticed that there are some unsuccessful runs in \denote{Exp}{Ryd\'en}, where the instrument got stuck in certain regions after receiving false guidance because of noisy surface point clouds, as illustrated in Fig.\ref{fig:Ryden, stuck in local regions}. Although in \denote{Exp}{Ryd{\'e}n} $d_{phantom} \leq d_{safe}=15$ mm, suggesting that violations of safety regions are not acted upon, we speculate this is also due to noisy point clouds, leading to a smaller distance inquiry result between the instrument tip and its surrounding point cloud. From Fig.\ref{fig:Traj ttest}, the DIBS method demonstrates a higher tracking accuracy with statistical significance as the level of deformation increases, which matches our previous finding in \ref{subsec:static_tracking}.

\begin{figure}[t]
\centering
\includegraphics[scale=0.42]{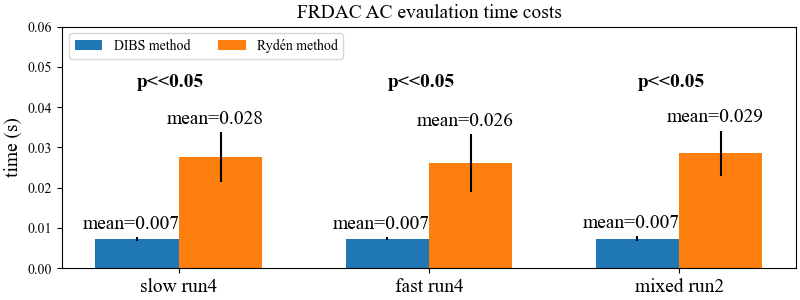}
\caption{Comparison of AC evaluation time between \denote{Exp}{ours} and \denote{Exp}{Ryd{\'e}n} in several runs.}
\label{fig:AC_evaluation_time}
\end{figure}

The overall tracking time recorded from TABLE \ref{tab:time costs DIBS} and \ref{tab:time costs Ryden} shows that the DIBS method is more time-efficient although it involves point cloud registration. We also find that although some runs in \denote{Exp}{Ryd{\'e}n}, for example, ``Slow run4", ``Fast run4" and ``Mixed run2", contain less points on the needle tip trajectory than their counterpart in \denote{Exp}{ours}, it took more time to complete a full tracking run. We analyzed the time spent on AC evaluation using both FRDAC methods for these trials, and the results are presented in Fig.\ref{fig:AC_evaluation_time}. The results show that the time cost for AC evaluation using the DIBS method is more consistent, whereas Ryd{\'e}n's method can be more time consuming, requiring more iterations before the proxy status reaches convergence, as shown in Fig.\ref{fig:AC evaluation, Ryden}.

To evaluate the potential of our FRDAC method for use in a modern clinical setting, we summarized the computational cost incurred for the two methods, in TABLE.\ref{tab:compare}. Our method, even though it involves point cloud registration, is still capable of running consistently at $43.48$ Hz without external GPU acceleration, which is faster than the depth capturing rate of the camera and the tissue deformation rate achieved in these experiments. This confirms that our method is fast enough to allow us to respond to tissue deformation changes appropriately, within the context of these experiments. On average, our pipeline runs even faster than Ryd\'en's method, which sometimes requires more time for AC evaluation.

\begin{table}
    \begin{center}
    \begin{tabular}{| c | c | c |}
    \hline
     Methods & Ours & Ryd{\'e}n's \\ \hline
     FPFH extraction time & 0.028s (initial round only) & None \\ \hline
     Registration time & 0.016s & None \\ \hline 
     Registration size & (8,000, 8,000) & None \\ \hline 
     AC evaluation size & 122,806 polygons & 1,000 points \\ \hline 
     \makecell{mean AC evaluation time} & 0.007s & 0.028s\\ \hline
     Total time cost & 0.023s & 0.028s\\ \hline
     Frequency & 43.48Hz & 35.71Hz \\ \hline
    \end{tabular}
    \caption{Comparison of the computational cost of the FRDAC developed by us and Ryd{\'e}n \textit{et al.}\cite{Ryden_proxy}}
    \label{tab:compare}
    \end{center}
\end{table}

Although the accuracy of the DIBS method is not directly affected by a local noisy point cloud, it relies on the global affine registration performance. Therefore, refining surface point may lead to better performance not only for the DIBS method but also for Ryd\'en's method, which fails to extract point clouds surrounding the instrument in situations where, for instance, occlusions are present. Furthermore, to accommodate different types of deformation, for example, sudden tool-tissue interactions, non-rigid registration algorithms should be considered.



\begin{figure}[t!]
    \centering

    \begin{subfigure}[t]{\columnwidth}
        \centering
        \includegraphics[width=\textwidth]{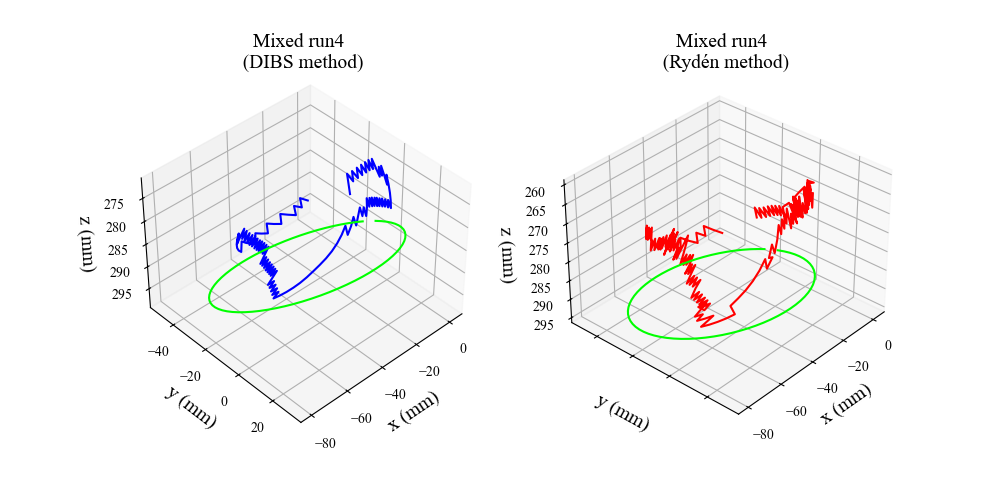}
    \end{subfigure}
    
    \caption{Comparison of dynamic tracking (Mixed run 4). Green, blue and red represent the reference trajectory, the actual trajectory in \denote{Exp}{ours} and in \denote{Exp}{Ryd\'en}, respectively.}
    \label{fig:Dynamic Tracking, mixed run4}
\end{figure}

\begin{figure}[t!]
    \centering
    \begin{subfigure}[t]{0.45\columnwidth}
        \centering
        \includegraphics[width=\textwidth]{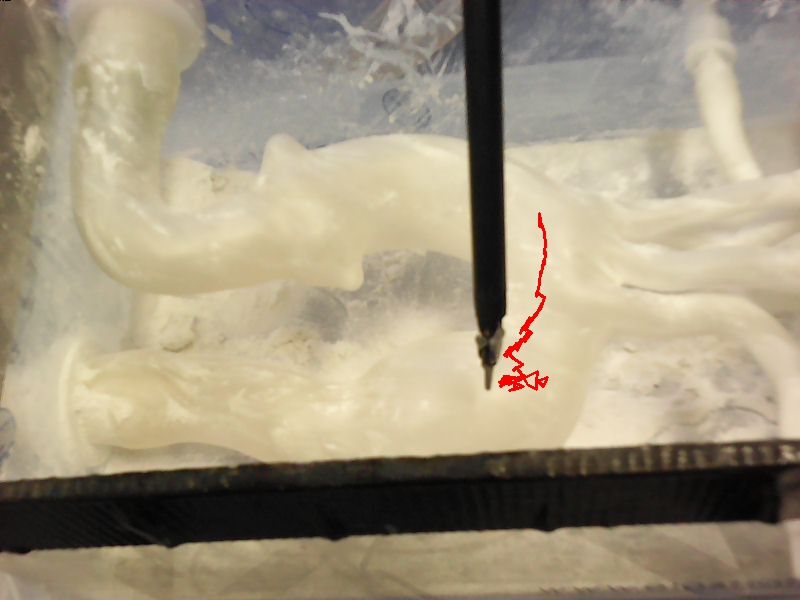}
        \caption{Unsuccessful run, frame 1}
    \end{subfigure}%
    ~ 
    \begin{subfigure}[t]{0.45\columnwidth}
        \centering
        \includegraphics[width=\textwidth]{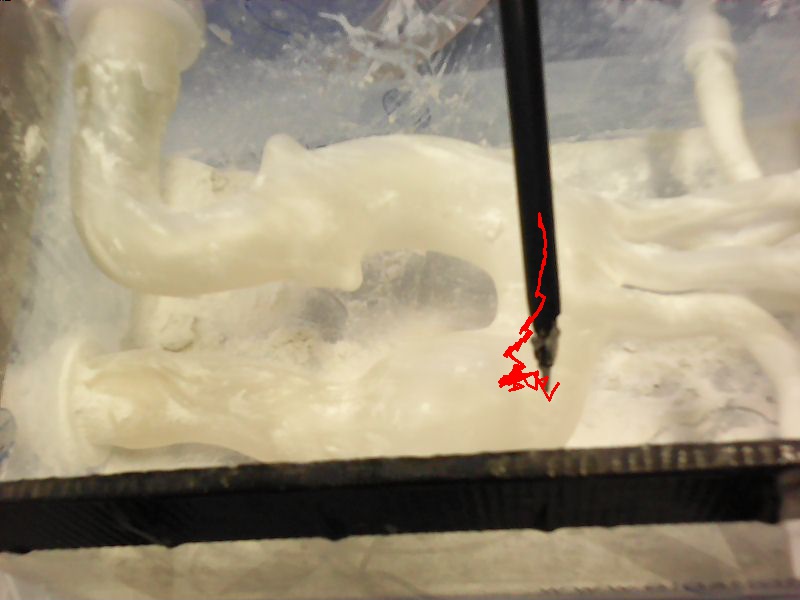}
        \caption{Unsuccessful run, frame 2}
    \end{subfigure}
    ~
    \begin{subfigure}[t]{0.45\columnwidth}
        \centering
        \includegraphics[width=\textwidth]{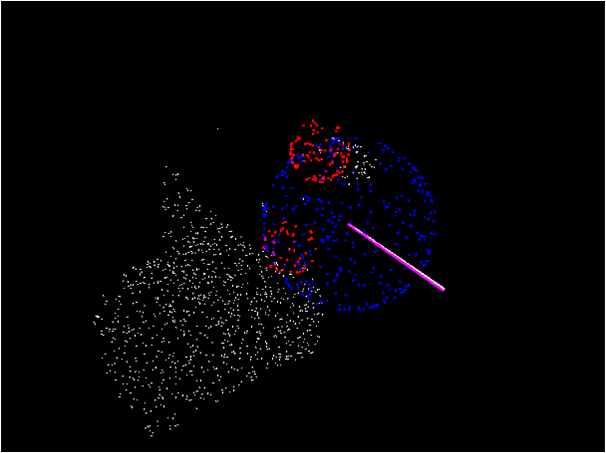}
        \caption{AC enforcement, frame 1}
    \end{subfigure}%
    ~ 
    \begin{subfigure}[t]{0.45\columnwidth}
        \centering
        \includegraphics[width=\textwidth]{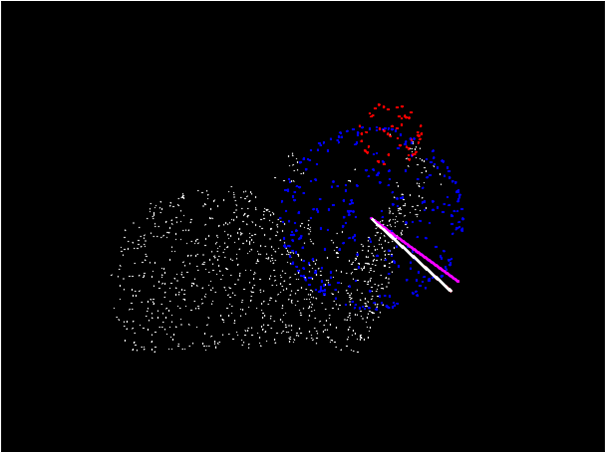}
        \caption{AC enforcement, frame 2}
    \end{subfigure}
    \caption{Illustration of AC enforcement in \denote{Exp}{Ryd\'en} where Ryd\'en's method generates false guidance because of noisy point cloud. Red spherical point clouds represent points in entrenched state. The white vector represents \vecdenote{v}{pen}, and the violet vector represents \vecdenote{v}{mov}. In frame 1 and 2, \vecdenote{v}{pen} does not reflect the actual surface normal direction, which leads to \vecdenote{v}{mov} drifting away from the reference trajectory.}
    \label{fig:Ryden, stuck in local regions}
\end{figure}

\begin{figure}[t!]
    \centering
    \begin{subfigure}[t]{0.45\columnwidth}
        \centering
        \includegraphics[width=\textwidth]{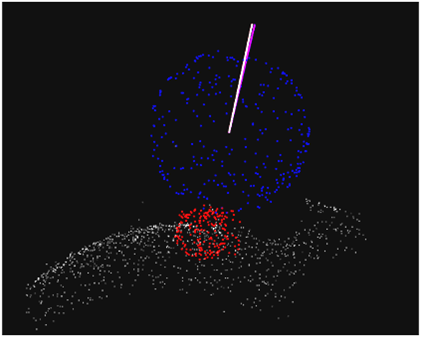}
        \caption{Case 1, AC evaluation time=0.0003s.}
    \end{subfigure}%
    ~ 
    \begin{subfigure}[t]{0.45\columnwidth}
        \centering
        \includegraphics[width=\textwidth]{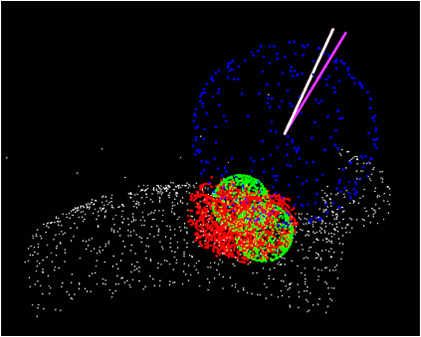}
        \caption{Case 2, AC evaluation time=0.0644s.}
    \end{subfigure}
    \caption{AC evaluation in \denote{Exp}{Ryd\'en} between the tool and the downsampled surface point cloud, with a size of 1,000.  No surface points are within the detection range of the current HIP. Red \add{and green} point clouds represent surface points that are within the detection range of the current HIP.}
    \label{fig:AC evaluation, Ryden}
\end{figure}

\begin{table}[]
\centering
\begin{tabular}{|l|l|l|l|l|}
\hline
\makecell{Run No.} & \makecell{1st} & \makecell{2nd} & \makecell{3rd} & \makecell{4th} \\ \hline
\makecell{\textbf{Slow} run time}& \makecell{7min 22s} & \makecell{7min 52s} & \makecell{7min 57s} & \makecell{7min 56s} \\ \hline
\makecell{\textbf{Slow} $n_{points}$}& \makecell{198} & \makecell{213} & \makecell{211} & \makecell{215} \\ \hline
\makecell{\textbf{Fast} run time}& \makecell{4min 19s} & \makecell{4min 19s} & \makecell{4min 36s} & \makecell{4min 47s} \\ \hline
\makecell{\textbf{Fast} $n_{points}$}& \makecell{117} & \makecell{119} & \makecell{125} & \makecell{130} \\ \hline
\makecell{\textbf{Mixed} run time}& \makecell{4min 31s} & \makecell{4min 30s} & \makecell{5min 8s} & \makecell{4min 29s} \\ \hline
\makecell{\textbf{Mixed} $n_{points}$}& \makecell{120} & \makecell{120} & \makecell{137} & \makecell{120} \\ \hline
\end{tabular}
\caption{Dynamic trajectory tracking time in \denote{Exp}{ours}}
\label{tab:time costs DIBS}
\end{table}

\begin{table}[]
\centering
\begin{tabular}{|l|l|l|l|l|}
\hline
\makecell{Run No.} & \makecell{1st} & \makecell{2nd} & \makecell{3rd} & \makecell{4th} \\ \hline
\makecell{\textbf{Slow} run time}& \makecell{9min 59s} & \makecell{8min 16s} & \makecell{7min 25s} & \makecell{9min 40s} \\ \hline
\makecell{\textbf{Slow} $n_{points}$}& \makecell{218} & \makecell{176} & \makecell{160} & \makecell{211} \\ \hline
\makecell{\textbf{Fast} run time}& \makecell{6min 25s} & \makecell{7min 28s} & \makecell{5min 16s} & \makecell{8min 17s} \\ \hline
\makecell{\textbf{Fast} $n_{points}$}& \makecell{141} & \makecell{165} & \makecell{116} & \makecell{183} \\ \hline
\makecell{\textbf{Mixed} run time}& \makecell{5min 29s} & \makecell{5min} & \makecell{9min 59s} & \makecell{5min} \\ \hline
\makecell{\textbf{Mixed} $n_{points}$}& \makecell{118} & \makecell{113} & \makecell{219} & \makecell{115} \\ \hline
\end{tabular}
\caption{Dynamic trajectory tracking time in \denote{Exp}{Ryd\'en}}
\label{tab:time costs Ryden}
\end{table}

\section{Conclusion} \label{sec:conclusion}
In this study, we present the pipeline design for a dissipative FRDAC controller that utilizes a depth-sensing camera to accommodate dynamic tissue deformation. An \textit{a priori} phantom model is leveraged to register with a streamed point cloud on the deforming tissue surface, updating at interactive rates, thereby enabling an always current mesh-represented AC at a speed that allows interactive guidance within a clinical environment. \textit{In vitro} experiments were conducted to verify the effectiveness of the proposed FRDAC pipeline in assisting a teleoperative trajectory tracking task implemented by a ``virtual surgeon". Our experimental results have indicated the capability of our FRDAC controller to fulfill the task safety requirement at a millimeter level, and we also provided quantitative analysis of the advantage of our mesh-represented FRDAC over the state-of-the-art, a sphere-represented FRDAC strategy. As a proof of concept implementation, the human element was removed in this study. Future work will involve conducting user studies to examine the effectiveness of our method in a real teleoperative surgical scenario. 

\addtolength{\textheight}{-0cm}
\bibliographystyle{ieeetr}
\bibliography{Files/references} 

@ARTICLE{Berkeley_Automation,
  author={Hwang, Minho and Thananjeyan, Brijen and Paradis, Samuel and Seita, Daniel and Ichnowski, Jeffrey and Fer, Danyal and Low, Thomas and Goldberg, Ken},
  journal={IEEE Robotics and Automation Letters}, 
  title={Efficiently Calibrating Cable-Driven Surgical Robots With RGBD Fiducial Sensing and Recurrent Neural Networks}, 
  year={2020},
  volume={5},
  number={4},
  pages={5937-5944},
  doi={10.1109/LRA.2020.3010746}}

@article{DIBS,
author = {Stuart A Bowyer and Ferdinando Rodriguez y Baena},
title ={Deformation invariant bounding spheres for dynamic active constraints in surgery},
journal = {Proceedings of the Institution of Mechanical Engineers, Part H: Journal of Engineering in Medicine},
volume = {228},
number = {4},
pages = {350-361},
year = {2014},
doi = {10.1177/0954411914527440}}

@INPROCEEDINGS{1000HZ_requirement,
  author={Zhuang, Y. and Canny, J.},
  booktitle={Proceedings 2000 ICRA. Millennium Conference. IEEE International Conference on Robotics and Automation. Symposia Proceedings (Cat. No.00CH37065)}, 
  title={Haptic interaction with global deformations}, 
  year={2000},
  volume={3},
  number={},
  pages={2428-2433 vol.3},
  doi={10.1109/ROBOT.2000.846391}}

@ARTICLE{Stuart_review,
  author={Bowyer, Stuart A. and Davies, Brian L. and Rodriguez y Baena, Ferdinando},
  journal={IEEE Transactions on Robotics}, 
  title={Active Constraints/Virtual Fixtures: A Survey}, 
  year={2014},
  volume={30},
  number={1},
  pages={138-157},
  doi={10.1109/TRO.2013.2283410}}

@article{rosenberg_use_1992,
  title={The Use of Virtual Fixtures As Perceptual Overlays to Enhance Operator Performance in Remote Environments: Interim Report for The Period June 1992 to July 1992},
  author={Rosenberg, LB},
  journal={Wright-Patterson Air Force Base, Ohio},
  year={1992}
}

@INPROCEEDINGS{Rosenberg_1993,
  author={Rosenberg, L.B.},
  booktitle={Proceedings of IEEE Virtual Reality Annual International Symposium}, 
  title={Virtual fixtures: Perceptual tools for telerobotic manipulation}, 
  year={1993},
  volume={},
  number={},
  pages={76-82},
  doi={10.1109/VRAIS.1993.380795}}

@INPROCEEDINGS{Ryden_proxy,
  author={Rydén, Fredrik and Chizeck, Howard Jay},
  booktitle={2012 IEEE/RSJ International Conference on Intelligent Robots and Systems}, 
  title={Forbidden-region virtual fixtures from streaming point clouds: Remotely touching and protecting a beating heart}, 
  year={2012},
  volume={},
  number={},
  pages={3308-3313},
  doi={10.1109/IROS.2012.6386012}}

@article{Ryden_journal,
author = {Sina Nia Kosari and Fredrik Rydén and Thomas S. Lendvay and Blake Hannaford and Howard Jay Chizeck},
title = {Forbidden region virtual fixtures from streaming point clouds},
journal = {Advanced Robotics},
volume = {28},
number = {22},
pages = {1507-1518},
year  = {2014},
publisher = {Taylor & Francis},
doi = {10.1080/01691864.2014.962613},

URL = {https://doi.org/10.1080/01691864.2014.962613}
}

@INPROCEEDINGS{Ryden_proxy_rudimentary,
  author={Rydén, Fredrik and Nia Kosari, Sina and Chizeck, Howard Jay},
  booktitle={2011 IEEE/RSJ International Conference on Intelligent Robots and Systems}, 
  title={Proxy method for fast haptic rendering from time varying point clouds}, 
  year={2011},
  volume={},
  number={},
  pages={2614-2619},
  doi={10.1109/IROS.2011.6094673}}

@article{CPD,
  title={Point set registration: Coherent point drift},
  author={Myronenko, Andriy and Song, Xubo},
  journal={IEEE transactions on pattern analysis and machine intelligence},
  volume={32},
  number={12},
  pages={2262--2275},
  year={2010},
  publisher={IEEE}
}

@article{BCPD,
  title={A Bayesian formulation of coherent point drift},
  author={Hirose, Osamu},
  journal={IEEE transactions on pattern analysis and machine intelligence},
  volume={43},
  number={7},
  pages={2269--2286},
  year={2020},
  publisher={IEEE}
}

@article{BCPD++,
  title={Acceleration of non-rigid point set registration with downsampling and Gaussian process regression},
  author={Hirose, Osamu},
  journal={IEEE Transactions on Pattern Analysis and Machine Intelligence},
  volume={43},
  number={8},
  pages={2858--2865},
  year={2020},
  publisher={IEEE}
}

@article{hybrid_force_position1981,
  title={Hybrid position/force control of manipulators},
  author={Raibert, Marc H and Craig, John J},
  year={1981}
}

@ARTICLE{AC_Acrobot,
  author={Davies, B. and Jakopec, M. and Harris, S.J. and Rodriguez Y Baena, F. and Barrett, A. and Evangelidis, A. and Gomes, P. and Henckel, J. and Cobb, J.},
  journal={Proceedings of the IEEE}, 
  title={Active-Constraint Robotics for Surgery}, 
  year={2006},
  volume={94},
  number={9},
  pages={1696-1704},
  doi={10.1109/JPROC.2006.880680}}

@article{daVinci_haptic_limitation,
  title={Technical review of the da Vinci surgical telemanipulator},
  author={Freschi, Cinzia and Ferrari, Vincenzo and Melfi, Franca and Ferrari, Mauro and Mosca, Franco and Cuschieri, Alfred},
  journal={The International Journal of Medical Robotics and Computer Assisted Surgery},
  volume={9},
  number={4},
  pages={396--406},
  year={2013},
  publisher={Wiley Online Library}
}

@article{Yang20tro-teaser,
  title={{TEASER: Fast and Certifiable Point Cloud Registration}},
  author={H. Yang and J. Shi and L. Carlone},
  journal={{IEEE} Trans. Robotics},
  Year = {2020} 
}

@inproceedings{rusu2009FPFH,
  title={Fast point feature histograms (FPFH) for 3D registration},
  author={Rusu, Radu Bogdan and Blodow, Nico and Beetz, Michael},
  booktitle={2009 IEEE international conference on robotics and automation},
  pages={3212--3217},
  year={2009},
  organization={IEEE}
}

@INPROCEEDINGS{Stuart_Enforcement_ICRA,
  author={Bowyer, Stuart A. and Rodriguez y Baena, Ferdinando},
  booktitle={2014 IEEE International Conference on Robotics and Automation (ICRA)}, 
  title={Dynamic frictional constraints in translation and rotation}, 
  year={2014},
  volume={},
  number={},
  pages={2685-2692},
  doi={10.1109/ICRA.2014.6907244}}

@article{kikuuwe2008control,
  title={A control framework to generate nonenergy-storing virtual fixtures: Use of simulated plasticity},
  author={Kikuuwe, Ryo and Takesue, Naoyuki and Fujimoto, Hideo},
  journal={IEEE Transactions on Robotics},
  volume={24},
  number={4},
  pages={781--793},
  year={2008},
  publisher={IEEE}
}

@ARTICLE{StuartDynamicTRO,
  author={Bowyer, Stuart A. and Rodriguez y Baena, Ferdinando},
  journal={IEEE Transactions on Robotics}, 
  title={Dissipative Control for Physical Human–Robot Interaction}, 
  year={2015},
  volume={31},
  number={6},
  pages={1281-1293},
  doi={10.1109/TRO.2015.2477956}}

@inproceedings{dVRK_2014,
  title={An open-source research kit for the da Vinci{\textregistered} Surgical System},
  author={Kazanzides, Peter and Chen, Zihan and Deguet, Anton and Fischer, Gregory S and Taylor, Russell H and DiMaio, Simon P},
  booktitle={2014 IEEE international conference on robotics and automation (ICRA)},
  pages={6434--6439},
  year={2014},
  organization={IEEE}
}

@article{aorta_quantification,
  title={Quantification of respiratory movement of the aorta and side branches},
  author={Sailer, Anna M and Wagemans, Bart AJM and Das, Marco and de Haan, Michiel W and Nelemans, Patricia J and Wildberger, Joachim E and Schurink, Geert Willem H},
  journal={Journal of Endovascular Therapy},
  volume={22},
  number={6},
  pages={905--911},
  year={2015},
  publisher={SAGE Publications Sage CA: Los Angeles, CA}
}

@book{Blender,
 author = {Hess, Roland},
 title = {Blender Foundations: The Essential Guide to Learning Blender 2.6},
 year = {2010},
 isbn = {0240814304, 9780240814308},
 publisher = {Focal Press},
}

@article{marinho2019dynamic,
  title={Dynamic active constraints for surgical robots using vector-field inequalities},
  author={Marinho, Murilo Marques and Adorno, Bruno Vilhena and Harada, Kanako and Mitsuishi, Mamoru},
  journal={IEEE Transactions on Robotics},
  volume={35},
  number={5},
  pages={1166--1185},
  year={2019},
  publisher={IEEE}
}

@inproceedings{enayati2016dynamic,
  title={A dynamic non-energy-storing guidance constraint with motion redirection for robot-assisted surgery},
  author={Enayati, Nima and Costa, Eva C Alves and Ferrigno, Giancarlo and De Momi, Elena},
  booktitle={2016 IEEE/RSJ International Conference on Intelligent Robots and Systems (IROS)},
  pages={4311--4316},
  year={2016},
  organization={IEEE}
}

@article{ren2008dynamic,
  title={Dynamic 3-D virtual fixtures for minimally invasive beating heart procedures},
  author={Ren, Jing and Patel, Rajni V and McIsaac, Kenneth A and Guiraudon, Gerard and Peters, Terry M},
  journal={IEEE transactions on medical imaging},
  volume={27},
  number={8},
  pages={1061--1070},
  year={2008},
  publisher={IEEE}
}

@article{yamamoto2012augmented,
  title={Augmented reality and haptic interfaces for robot-assisted surgery},
  author={Yamamoto, Tomonori and Abolhassani, Niki and Jung, Sung and Okamura, Allison M and Judkins, Timothy N},
  journal={The International Journal of Medical Robotics and Computer Assisted Surgery},
  volume={8},
  number={1},
  pages={45--56},
  year={2012},
  publisher={Wiley Online Library}
}

@article{O3D,
    author    = {Qian-Yi Zhou and Jaesik Park and Vladlen Koltun},
    title     = {{Open3D}: {A} Modern Library for {3D} Data Processing},
    journal   = {arXiv:1801.09847},
    year      = {2018},
}

@Article{Matplotlib,
  Author    = {Hunter, J. D.},
  Title     = {Matplotlib: A 2D graphics environment},
  Journal   = {Computing in Science \& Engineering},
  Volume    = {9},
  Number    = {3},
  Pages     = {90--95},
  publisher = {IEEE COMPUTER SOC},
  doi       = {10.1109/MCSE.2007.55},
  year      = 2007
}
\end{document}